\documentclass[sigconf,nonacm]{acmart}

\usepackage{multirow}

\setcopyright{none}
\begin{document}

\title{WorkDrive: Roadwork Chain-of-Causation for Autonomous Driving}

\author{Tianyi Jiang}
\email{tianyijiang0219@gmail.com}
\affiliation{
  \institution{Peking University}
  \city{Beijing}
  \country{China}
}

\author{Wen Zhang}
\affiliation{
  \institution{Xiaomi EV}
  \city{Beijing}
  \country{China}
}

\author{Sihan Yang}
\affiliation{
  \institution{Peking University}
  \city{Beijing}
  \country{China}
}

\author{Ming Lu}
\affiliation{
  \institution{Peking University}
  \city{Beijing}
  \country{China}
}

\author{Wentao Zhang}
\authornote{Corresponding author.}
\email{wentao.zhang@pku.edu.cn}
\affiliation{
  \institution{Peking University}
  \city{Beijing}
  \country{China}
}

\renewcommand{\shortauthors}{Jiang et al.}

\begin{abstract}
Autonomous driving vision-language models (VLMs) struggle in roadwork zones, where familiar visual cues such as lane markings and permanent signs are altered or absent, and temporary devices such as cones and barriers redefine the drivable corridor. VLMs can detect these objects, but without explicit guidance they anchor their reasoning on familiar elements from pre-training and fail to connect work-zone observations to correct planning decisions. We propose WorkDrive, a framework that constructs perception-grounded causal reasoning for work zones and aligns it with trajectory prediction. An automated multitask perception pipeline extracts structured scene facts and injects them into a Chain-of-Causation (CoC) annotation pipeline, redirecting the annotator's attention to domain-specific elements. The resulting reasoning labels are used for supervised fine-tuning, followed by reinforcement learning with a single reward: consistency between lateral meta-actions and the predicted trajectory. On ROADWork, the largest public work-zone dataset, the proposed roadwork CoC reduces trajectory average displacement error (ADE) by 9.0\%, and consistency-based GRPO yields a further 3.0\%, achieving progressive improvement over the trajectory-only baseline. Code and data will be publicly released.
\end{abstract}



\maketitle

\section{Introduction}
Autonomous driving in structured road environments approaches practical maturity.
End-to-end frameworks now unify perception, prediction, and planning into a single differentiable pipeline~\cite{hu2023planning}, while vision-language models bring world knowledge and semantic understanding to the planning stage~\cite{hwang2024emma,jiang2024senna,zhou2026opendrivevla}.
Despite this progress, long-tail scenarios remain the central obstacle to safe real-world deployment: models learn almost exclusively from routine driving data, yet real-world accidents concentrate in rare situations that fall outside the training distribution.
Work zones are among the most critical of such scenarios: in the United States alone, approximately 100{,}000 work-zone crashes occur each year, resulting in over 40{,}000 injuries~\cite{penndot2019ads}.
What makes work zones particularly challenging is that the visual cues models learn to rely on in normal driving, such as lane markings, permanent signs, and stable road geometry, are altered or absent.
Instead, temporary traffic control devices such as cones and barriers redefine the drivable corridor, but their spatial arrangement varies from site to site and changes over time.
Models trained mainly on standard road layouts have no learned basis for interpreting these unfamiliar setups.
Correctly navigating a work zone therefore requires reasoning about what is currently visible in the scene rather than relying on patterns memorized from routine data.

\begin{figure}[t]
    \centering
    \includegraphics[width=\linewidth]{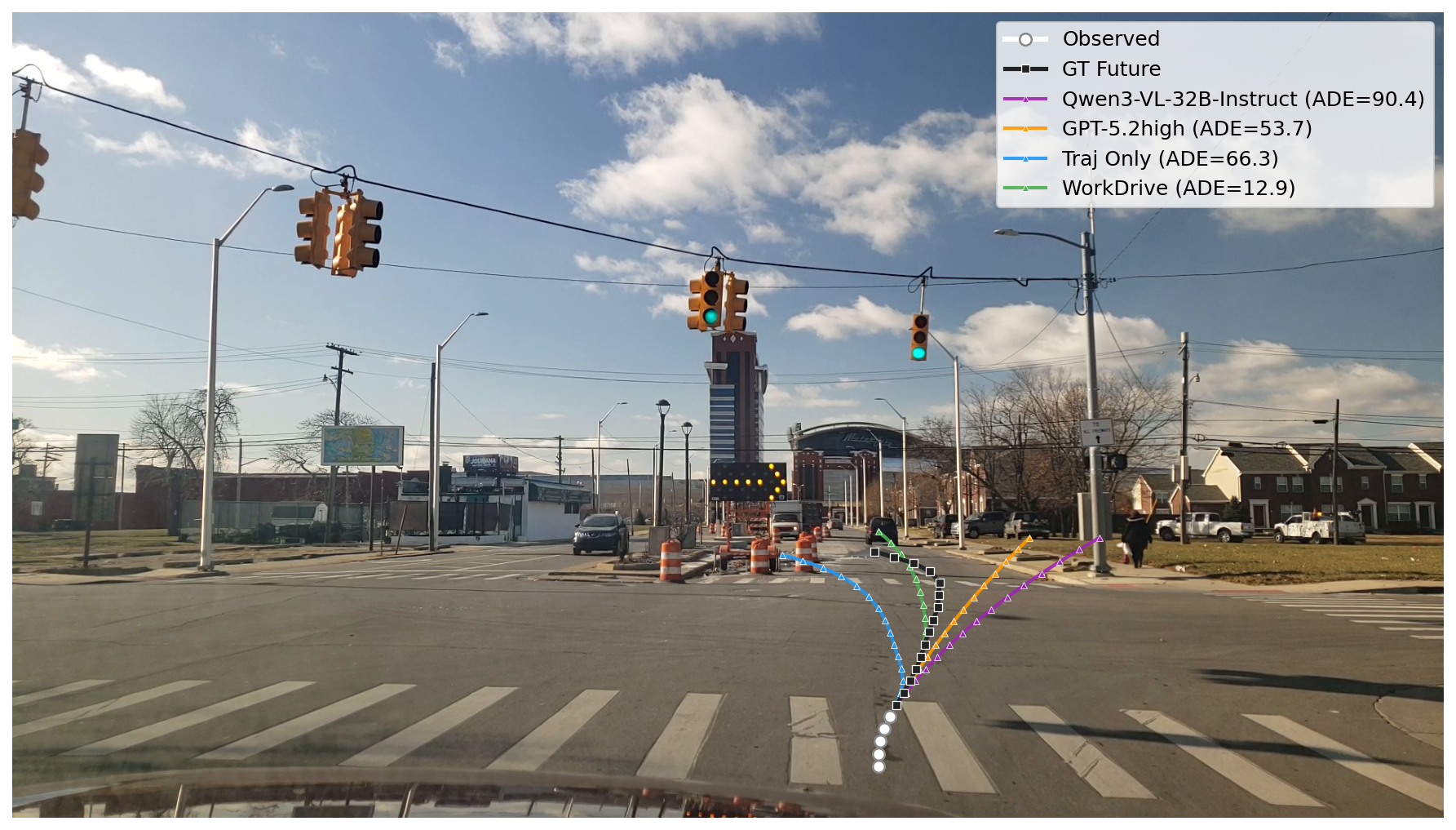}
    \caption{Trajectory prediction in a work-zone scene with lane closures and traffic cones. General-purpose VLMs and the trajectory-only SFT baseline all deviate significantly from the ground truth, while our \textbf{WorkDrive} produces the closest prediction, demonstrating the necessity of perception-grounded reasoning for OOD planning.}
    \label{fig:traj-case}
    \vspace{-0.1in}
\end{figure}

To address this challenge, recent research has progressively integrated reasoning into the planning pipeline.
Vision-language models~\cite{tian2024drivevlm,jiang2024senna,zhou2026opendrivevla,fu2025orion,jiang2025diffvla,peng2025colavla} introduce semantic understanding, yet a gap remains between high-level language descriptions and precise trajectory coordinates that a planner must produce.
Structured chain-of-thought reasoning narrows this gap by decomposing the planning task into interpretable reasoning steps before generating a trajectory, and several studies confirm that this decomposition improves both accuracy and safety~\cite{wang2024drivecot,sima2024drivelm,wang2025cot4ad}.
More recently, reinforcement learning has been used to tighten the link between reasoning and planning: methods such as AlphaDrive~\cite{jiang2025alphadrive} and Drive-R1~\cite{li2026drive} adopt Group Relative Policy Optimization (GRPO)~\cite{shao2024deepseekmath} and reward the consistency between reasoning-derived actions and predicted trajectories~\cite{zhang2025omnidrive,zheng2025driveagent}.
However, these methods derive their reasoning supervision from causal knowledge that large models already possess about routine driving, such as how lane markings and traffic signals relate to driving decisions.
In work zones, VLMs can detect temporary objects such as cones and barriers, but they lack the causal knowledge to connect these observations to correct planning decisions.
As illustrated in Figure~\ref{fig:traj-case}, even frontier VLMs perceive work-zone objects yet their trajectories deviate significantly from the ground truth, confirming that perception alone does not yield correct planning without domain-specific causal reasoning.

On the work-zone front, ROADWork~\cite{ghosh2025roadwork} provides the first large-scale work-zone driving dataset, but it contains only trajectory ground truth without reasoning annotations.
REACT-Drive~\cite{liao2025work} is the first to study VLM planning in work zones and proposes a retrieval-augmented generation strategy to improve trajectory prediction, yet it relies on matching predefined failure patterns and patching outputs with rules rather than strengthening the model's own reasoning ability.
The root cause is not that VLMs cannot see work-zone objects, but that they default to familiar reasoning patterns: without explicit guidance, a VLM anchors its reasoning on common elements encountered during pre-training, such as traffic lights and pedestrians, while overlooking the cones, barriers, and temporary signs that actually govern the drivable path.
Building correct causal reasoning for work zones therefore requires redirecting the model's attention to these domain-specific elements through structured perception information.

To bridge this gap, we propose \textbf{WorkDrive}, a perception-grounded reasoning annotation and consistency alignment framework for work-zone scenarios.
As illustrated in Figure~\ref{fig:framework}, the framework consists of three stages.
First, an automated multitask perception pipeline comprising depth estimation, open-vocabulary detection, instance segmentation, and cross-frame tracking extracts structured scene facts from raw ROADWork frames, providing the factual basis absent from existing work-zone datasets.
Second, building on the Chain-of-Causation framework of AR1~\cite{wang2025alpamayo}, an LLM-driven annotation pipeline consumes these perception outputs to generate causally grounded reasoning labels, which are then used for parameter-efficient fine-tuning of an 8B-parameter VLM via LoRA to inject both reasoning and planning capabilities.
Third, GRPO reinforcement learning employs the consistency between lateral meta-actions and the predicted trajectory as its sole reward signal, further aligning reasoning with planning.
Experiments show that perception-grounded reasoning annotations significantly improve work-zone trajectory planning, and that GRPO consistency alignment yields further gains over supervised fine-tuning using only a single consistency reward between the meta-action and the predicted trajectory.

Our contributions are as follows:

\begin{itemize}
    \item We propose a perception-grounded annotation framework that constructs causal reasoning data for work zones. An automated multitask perception pipeline extracts structured scene facts and injects them into the Chain-of-Causation annotation framework~\cite{wang2025alpamayo}, redirecting the annotator model's attention to domain-specific elements.
    \item We show that a single GRPO reward measuring the consistency between lateral meta-actions and the predicted trajectory suffices to improve planning over supervised fine-tuning, without any explicit trajectory reward.
    \item Perception-grounded reasoning reduces ADE by 9.0\%, and consistency-based GRPO adds a further 3.0\%, achieving progressive improvement. The code and reasoning dataset will be publicly released.
\end{itemize}
\section{Related Work}

\subsection{Vision-Language Models for Autonomous Driving}

Recent work integrates VLMs into autonomous driving at increasing depth.
Early frameworks use VLMs as scene describers that feed hierarchical planners~\cite{xu2024drivegpt4,tian2024drivevlm}; more recent vision-language-action architectures directly regress trajectory waypoints from visual inputs~\cite{jiang2024senna,hwang2024emma,zhou2026opendrivevla,fu2025orion,jiang2025diffvla}, with CoLaVLA~\cite{peng2025colavla} further strengthening multi-view spatial understanding through collaborative layout awareness.
These architectures establish a direct vision-to-trajectory mapping, but the gap between language-level understanding and precise coordinate prediction remains.
Chain-of-thought (CoT) reasoning narrows this gap by decomposing planning into interpretable steps.
DriveLM~\cite{sima2024drivelm} introduces graph-structured QA chains linking perception, prediction, and planning; DriveCoT~\cite{wang2024drivecot} adds a mark-then-reason pipeline that links reasoning tokens to visual regions; CoT4AD~\cite{wang2025cot4ad} and RecogDrive~\cite{li2025recogdrive} further explore multi-granularity reasoning and recognition-guided decomposition.
AR1~\cite{wang2025alpamayo} elevates CoT to causal explanation through its Chain-of-Causation (CoC) framework, which decomposes each decision into meta-action classification, causal factor identification, and a multi-hop chain tracing reasoning from perception to action.
However, supervised fine-tuning on reasoning tokens alone does not guarantee that the generated rationale is consistent with the output trajectory.
Reinforcement learning addresses this: AlphaDrive~\cite{jiang2025alphadrive} first applies GRPO~\cite{shao2024deepseekmath} to driving VLMs with format and trajectory rewards; subsequent work extends the paradigm with adaptive reasoning depth~\cite{luo2025adathinkdrive,zhou2025autovla}, multi-agent coordination~\cite{zheng2025driveagent,li2026drive}, and knowledge-distilled reward design~\cite{zhang2025omnidrive,zhang2025openread}.
This line of work establishes a two-stage SFT-then-RL training paradigm.

These methods share two limitations.
First, all existing reasoning labels are generated by the VLM itself or composed from manual templates, without grounding in external perception models; whether the rationale faithfully reflects the physical scene is neither guaranteed nor verifiable.
Second, both reasoning supervision and RL alignment have only been validated on routine datasets such as nuScenes~\cite{caesar2020nuscenes} and Waymo~\cite{sun2020scalability}, leaving out-of-distribution long-tail scenarios untested.
We address both by anchoring reasoning in an automated perception pipeline and validating consistency-based RL on OOD work-zone data.

\begin{figure*}[t]
    \centering
    \includegraphics[width=\textwidth]{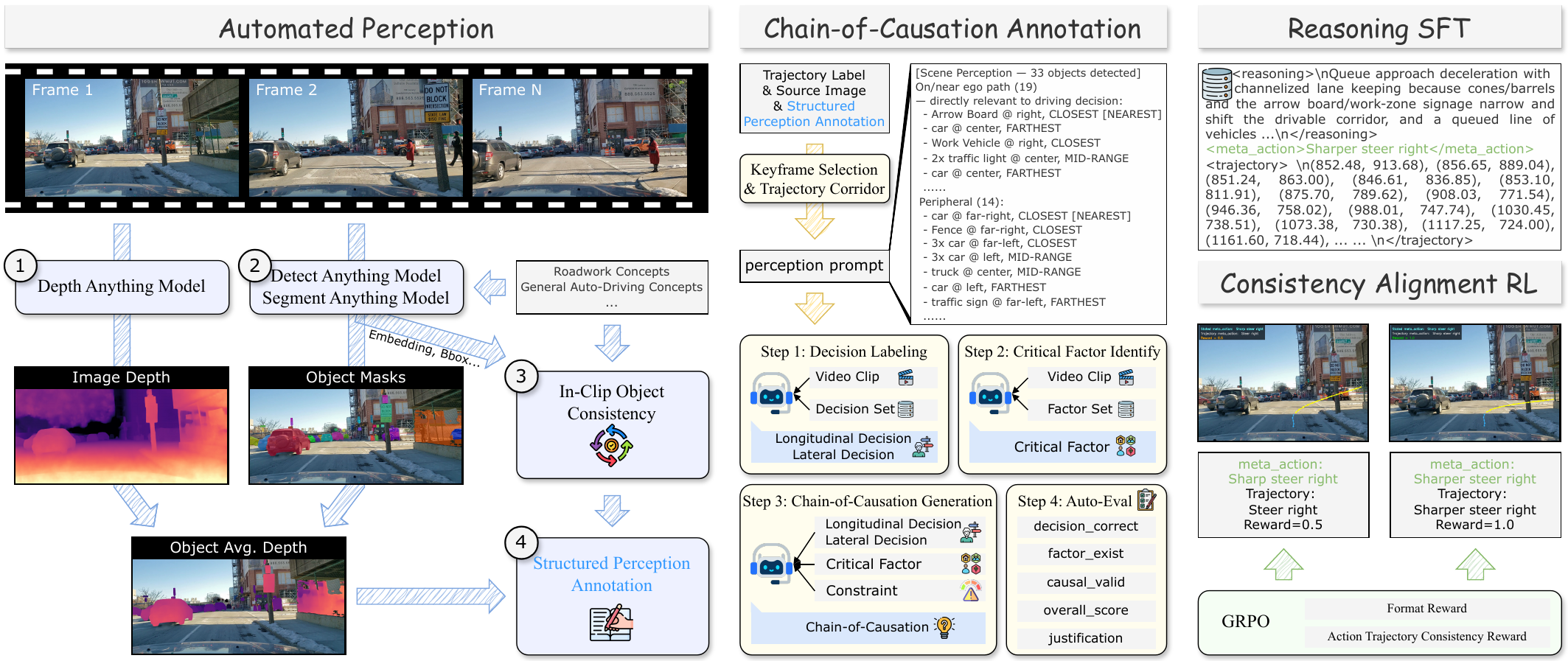}
    \caption{Overview of the WorkDrive framework. \textbf{Phase~I} extracts structured scene representations via depth estimation, open-vocabulary detection, instance segmentation, and cross-frame tracking. \textbf{Phase~II} generates Chain-of-Causation reasoning annotations through decision classification, causal factor identification, causal chain synthesis, and quality assessment, all grounded in Phase~I outputs. \textbf{Phase~III} performs LoRA-based supervised fine-tuning followed by GRPO reinforcement learning with a meta-action consistency reward.}
    \label{fig:framework}
\end{figure*}

\subsection{Work Zones in Autonomous Driving}

Work zones dynamically reshape the drivable space through temporary traffic control devices such as cones, barrels, and arrow boards, causing the structured priors of routine driving to fail~\cite{mutcd}, as evidenced by AV incidents in construction areas~\cite{tesla2017,cruise2023}.
On the perception side, early work focuses on temporary lane detection~\cite{gumpp2009recognition,mathibela2013roadwork}; subsequent datasets address corner-case and semantic segmentation~\cite{li2022coda,kim2024rosa,sural2026workzone3d}.
ROADWork~\cite{ghosh2025roadwork} provides the first large-scale work-zone benchmark with 4{,}375 videos across 18 cities, covering lane-level attributes, work-zone activity types, and trajectory ground truth, but does not include reasoning annotations.
On the planning side, an operational safety framework has been proposed for work-zone navigation~\cite{sahu2025towards}.
REACT-Drive~\cite{liao2025work} is the first to study VLM-based planning in work zones; it diagnoses systematic failures of baseline VLMs and proposes retrieval-augmented generation (RAG) to improve trajectory prediction by matching observed failure patterns to a predefined knowledge base and patching the output accordingly.
While effective as a post-processing strategy, RAG corrects planning outputs without strengthening the model's own reasoning about work-zone elements.

The causal reasoning layer connecting work-zone perception to planning therefore remains absent: existing datasets provide trajectory ground truth but not structured decision rationales, and existing planning methods correct outputs without building the model's capacity to reason about domain-specific elements.
We construct perception-anchored Chain-of-Causation annotations on ROADWork and inject them into a VLM through SFT and consistency-based RL, validating perception-grounded reasoning in work-zone scenarios for the first time.
\section{Method}

\subsection{Framework Overview}

Given a video clip $\mathcal{V} = \{I_1, I_2, \ldots, I_T\}$ from the ROADWork dataset~\cite{ghosh2025roadwork} and the corresponding ego-vehicle trajectory $\tau = \{(x_t, y_t)\}_{t=1}^{T}$, our goal is to train a vision-language model $f_\theta$ that, at a keyframe $I_k$, receives the keyframe together with two preceding history frames $\mathcal{I}_k = \{I_{k-2}, I_{k-1}, I_k\}$ and the observed trajectory $\tau_{\text{obs}}$, and sequentially produces causal reasoning $\mathcal{R}$, a lateral meta-action $\mathcal{M}$, and a predicted trajectory $\hat{\tau}_{\text{pred}}$:
\begin{equation}
    (\mathcal{R},\; \mathcal{M},\; \hat{\tau}_{\text{pred}}) = f_\theta(\mathcal{I}_k,\; \tau_{\text{obs}}).
    \label{eq:reasoning-model}
\end{equation}
Here $\mathcal{R}$ denotes Chain-of-Causation~\cite{wang2025alpamayo} reasoning text, $\mathcal{M} \in \{\textit{Go straight},\; \textit{Steer left},\; \textit{Steer right},\; \textit{Sharp steer left},\; \textit{Sharp steer right}\}$ is the lateral intent derived from the reasoning, and $\hat{\tau}_{\text{pred}}$ is a future trajectory of 15 waypoints in pixel coordinates following ROADWork's trajectory format.
The frame interval is fixed at 0.4\,s; $\tau_{\text{obs}}$ consists of 5 waypoints up to and including the keyframe position, providing the model with recent ego-motion context.
By contrast, the baseline model directly outputs $\hat{\tau}_{\text{pred}} = f_\theta(\mathcal{I}_k, \tau_{\text{obs}})$ without reasoning or meta-action.
The difference between the two formulations embodies the core hypothesis of this work: in out-of-distribution work-zone scenarios, grounding the reasoning process in structured perception of domain-specific elements and enforcing consistency between reasoning and trajectory yield better planning than learning from trajectory supervision alone.

The framework consists of three phases (Figure~\ref{fig:framework}). \textbf{Phase~I} (\S\ref{sec:perception}) extracts structured scene representations from raw frames via an automated perception pipeline. \textbf{Phase~II} (\S\ref{sec:coc}) consumes these representations to generate CoC reasoning annotations through a four-step causally-constrained pipeline. \textbf{Phase~III} (\S\ref{sec:alignment}) first performs LoRA~\cite{hu2022lora}-based SFT to inject reasoning and planning capabilities, then applies GRPO~\cite{shao2024deepseekmath} alignment with meta-action consistency as the sole reward.

\subsection{Automated Perception Pipeline}
\label{sec:perception}

When generating causal explanations, VLMs tend to anchor on familiar elements from pre-training rather than the work-zone objects that actually govern the drivable path.
To counter this, we decouple fact extraction from causal reasoning: specialized vision models produce structured scene representations offline, and these are injected as mandatory context into the downstream annotation pipeline, redirecting the annotator's attention to domain-specific elements.
This pipeline comprises four stages and produces a per-frame structured JSON without manual annotation.

\textbf{Monocular depth estimation.}
Depth~Anything~3 (DA3-Giant)~\cite{lin2025depth} outputs an affine-invariant relative depth map for each frame at $1920 \times 1080$ resolution.
Because the physical scale of these maps is not comparable across frames, we compute per-frame percentile thresholds at $33\%$ and $66\%$ of all detected objects' mean depth values as division points and categorize objects into three depth levels $\{\textsc{closest},\; \textsc{mid-range},\; \textsc{farthest}\}$, corresponding to the bands $\leq P_{33}$, $(P_{33},\, P_{66}]$, and $> P_{66}$ respectively.

\textbf{Open-vocabulary object detection.}
Rex-Omni~\cite{jiang2025detect} performs two-pass recognition: the first pass covers 45 work-zone-specific categories from the ROADWork~\cite{ghosh2025roadwork} taxonomy; the second covers 11 general traffic categories from nuScenes~\cite{caesar2020nuscenes} and Waymo~\cite{sun2020scalability} (full lists in Appendix~A).
When the two passes overlap (IoU $> 0.5$), the work-zone category is retained, because work-zone objects carry higher decision-making salience.

\textbf{Instance segmentation and embedding extraction.}
For each detected bounding box $b_i$ in frame $I_t$, SAM~3~\cite{carion2025sam} receives $b_i$ as a box prompt and outputs an instance mask $m_i$ together with a feature embedding $\mathbf{e}_i \in \mathbb{R}^{d}$:
\begin{equation}
    (m_i,\; \mathbf{e}_i) = \mathrm{SAM3}(I_t,\; b_i).
    \label{eq:sam3}
\end{equation}
When multiple candidates are returned, the highest-IoU mask is selected.
The embeddings $\{\mathbf{e}_i\}$ are carried forward for cross-frame instance matching.

\textbf{Cross-frame tracking.}
An instance $i$ in frame~$t$ is linked to instance $j$ in frame~$t{+}1$ if both spatial overlap and appearance similarity exceed their respective thresholds:
\begin{equation}
    \mathrm{IoU}(b_i^{t},\, b_j^{t+1}) > 0.3 \;\wedge\; \frac{\mathbf{e}_i^{t} \cdot \mathbf{e}_j^{t+1}}{\|\mathbf{e}_i^{t}\|\,\|\mathbf{e}_j^{t+1}\|} > 0.7.
    \label{eq:tracking}
\end{equation}
When class labels within a track are inconsistent, majority voting unifies them to ensure cross-frame conceptual consistency.

\textbf{Perception information aggregation.}
All outputs are assembled into a per-frame JSON; each entry records its category, bounding box, instance mask, depth level, and track ID.

\begin{figure}
    \centering
    \includegraphics[width=0.85\columnwidth]{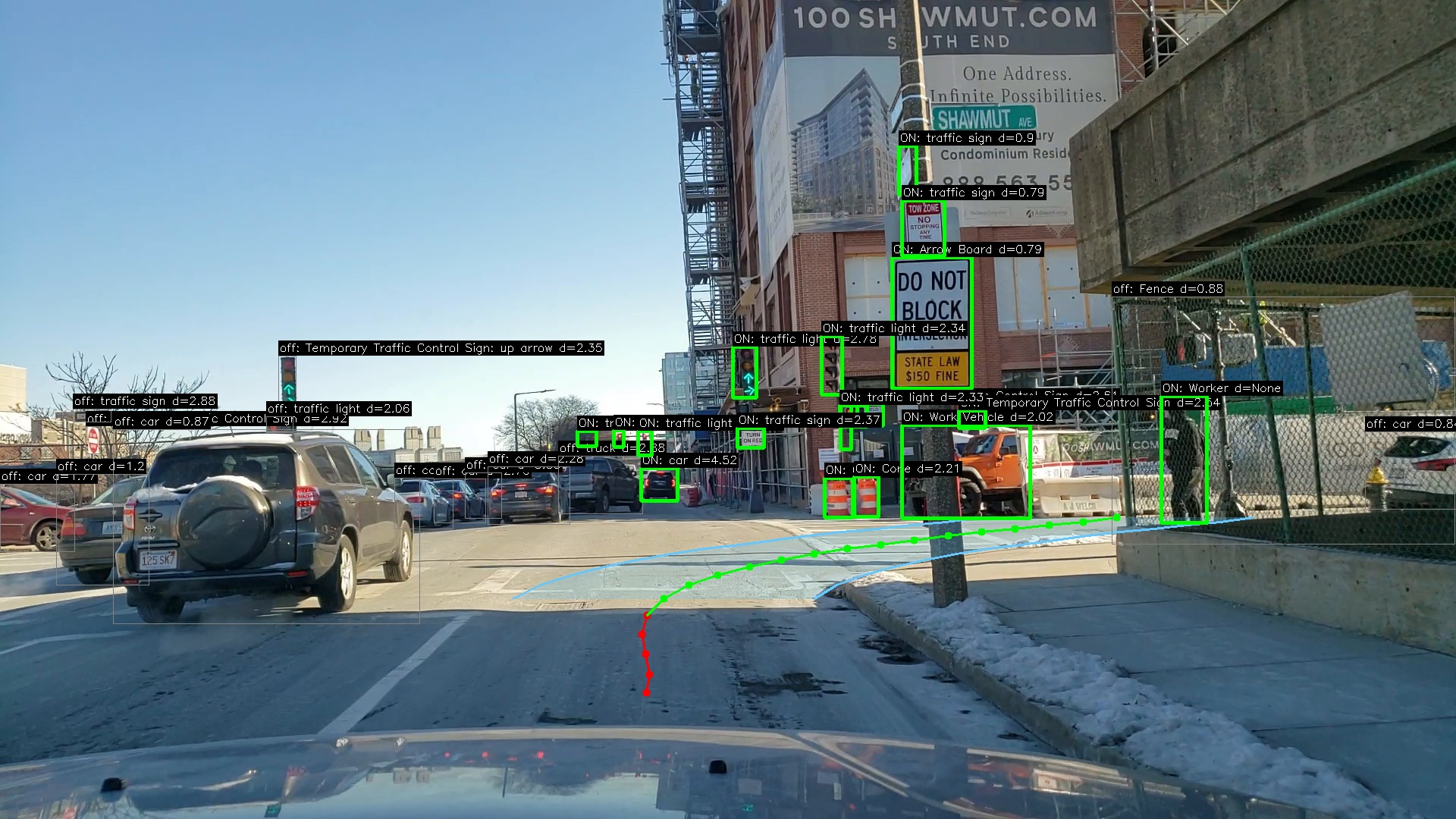}
    \caption{Driving corridor and on-path classification.  The cyan dashed boundary delineates the corridor around the future trajectory (green dots).  Objects whose bounding-box bottom-center falls inside the corridor are labeled \textbf{ON-PATH} (bright green); others are \textbf{peripheral} (dark labels).}
    \label{fig:corridor}
\end{figure}

\subsection{Chain-of-Causation Annotation Pipeline}
\label{sec:coc}

Phase~I provides structured scene facts for each frame; Phase~II builds on these to answer a deeper question: \emph{why did the driver make this decision?}
We decompose CoC generation into four causally-constrained steps, each of which must receive Phase~I perception data as mandatory context to ensure that reasoning is grounded in observable scene evidence.

Before executing the generation steps, the per-frame structured perception information from Phase~I is rendered into a perception prompt $\mathcal{P}_k$ that the annotator LLM can readily consume.
We design a two-tier object description scheme that separates \textbf{On-path objects}, those within the ego vehicle's driving corridor that directly influence decisions, from \textbf{Peripheral objects} that serve as contextual references.
Each object is annotated with its category, horizontal position across five equal bins, and depth level across three percentile tiers.
The driving corridor is defined based on the ego vehicle's future trajectory, with its half-width undergoing perspective scaling along the vertical image axis:
\begin{equation}
    w(y) = 60 + \frac{y}{H} \times 190 \quad \text{(pixels)},
    \label{eq:corridor}
\end{equation}
where $H$ is the image height.
The corridor is wider near the vehicle to accommodate lane-width margins and narrower at distance to match perspective foreshortening, as visualized in Figure~\ref{fig:corridor}.
Rather than performing pixel-level overlap on SAM~3 masks, we test only the bottom-center of each bounding box, which serves as a ground-contact proxy.
The test is $y$-driven: we interpolate the trajectory centerline $x$-coordinate at the query object's $y$-coordinate and classify the object as on-path whenever the horizontal offset falls within the corridor half-width:
\begin{equation}
    \bigl|\, x_{\text{obj}} - x_{\text{traj}}(y_{\text{obj}}) \,\bigr| \;\leq\; w(y_{\text{obj}}).
    \label{eq:corridor-test}
\end{equation}
This on-path/peripheral formatting directs the annotator LLM's attention toward decision-relevant objects during CoC generation, preventing dilution by numerous background targets.
Crucially, $\mathcal{P}_k$ is used \emph{only} during the annotation pipeline (Phase~II); at inference time the trained model receives only the keyframe, two preceding frames, and the observed trajectory, with no access to $\mathcal{P}_k$ or the GT future trajectory, thereby precluding any trajectory leakage from the annotation stage into model predictions.

\textbf{Step 1: Closed-set decision classification.}
The annotator LLM receives a six-frame window $\{I_{k-2},\ldots,I_{k+3}\}$ comprising two frames before the keyframe, the keyframe itself, and three frames after it, with observed and future trajectories rendered in red and green.
$\mathcal{P}_k$ is injected as a system prompt prefix, providing object counts and spatial distributions.
The model selects a longitudinal decision $d_{\text{lon}}$ (e.g., cruise, decelerate, stop) and a lateral decision $d_{\text{lat}}$ (e.g., lane keep, merge left) from 28 predefined categories (full list in Appendix~B), constrained to valid labels through structured-output enum enforcement:
\begin{equation}
    (d_{\text{lon}},\, d_{\text{lat}}) = \mathrm{LLM}\!\bigl(\{I_{k-2},\ldots,I_{k+3}\},\; \mathcal{P}_k\bigr).
    \label{eq:step1}
\end{equation}

\textbf{Step 2: Causal factor identification.}
To \emph{prevent causal inversion}, the input is restricted to $\{I_{k-2},\, I_{k-1},\, I_k\}$, excluding future frames so that factors cannot be reverse-engineered from outcomes.
$\mathcal{P}_k$ requires the model to select causal sources from the on-path and peripheral entity lists.
A structured guidance framework covering ten major categories (Appendix~D) is provided as a prompt, while free-text descriptions are also permitted:
\begin{equation}
    \mathcal{F} = \mathrm{LLM}\!\bigl(\{I_{k-2},\, I_{k-1},\, I_k\},\; \mathcal{P}_k,\; d_{\text{lon}},\; d_{\text{lat}}\bigr).
    \label{eq:step2}
\end{equation}

\textbf{Step 3: Causal chain synthesis.}
The decision pair and factor set from the preceding two steps are synthesized into the causal reasoning chain $\mathcal{R}$.
Crucially, this step takes \emph{only text} as input, without any images, to prevent new visual information from altering already-established causal judgments.
$\mathcal{P}_k$ serves as a factual constraint, ensuring that the merged reasoning does not conflict with scene evidence.
$\mathcal{R}$ must follow the mandatory structure \texttt{[Decision] because [Key Factors]}, selecting the most proximate factors rather than enumerating all background conditions:
\begin{equation}
    \mathcal{R} = \mathrm{LLM}\!\bigl(d_{\text{lon}},\; d_{\text{lat}},\; \mathcal{F},\; \mathcal{P}_k\bigr).
    \label{eq:step3}
\end{equation}

\textbf{Step 4: Automated quality assessment.}
The annotation is reviewed against subsampled frames from the full clip $\mathcal{V}$, the only step with access to all frames.
$\mathcal{P}_k$ is used as factual grounding to detect hallucinated objects.
Assessment evaluates four criteria (causal coverage, causal correctness, proximate-cause adherence, and decision minimality) and produces a scalar quality score:
\begin{equation}
    s = \mathrm{LLM}\!\bigl(\mathcal{V},\; d_{\text{lon}},\; d_{\text{lat}},\; \mathcal{F},\; \mathcal{R},\; \mathcal{P}_k\bigr), \quad s \in [0,10].
    \label{eq:step4}
\end{equation}
Annotations are retained only when $s \geq 8$ and all core fields pass validation.
Steps~1--3 use either Qwen3-VL-32B-Instruct or GPT-5.2high as the annotator LLM; for Step~4, GPT-5.2high is selected as the sole judge based on its closest alignment with human ratings among three candidates (\S\ref{sec:human-eval}).

\begin{figure}[t]
    \centering
    \includegraphics[width=\columnwidth]{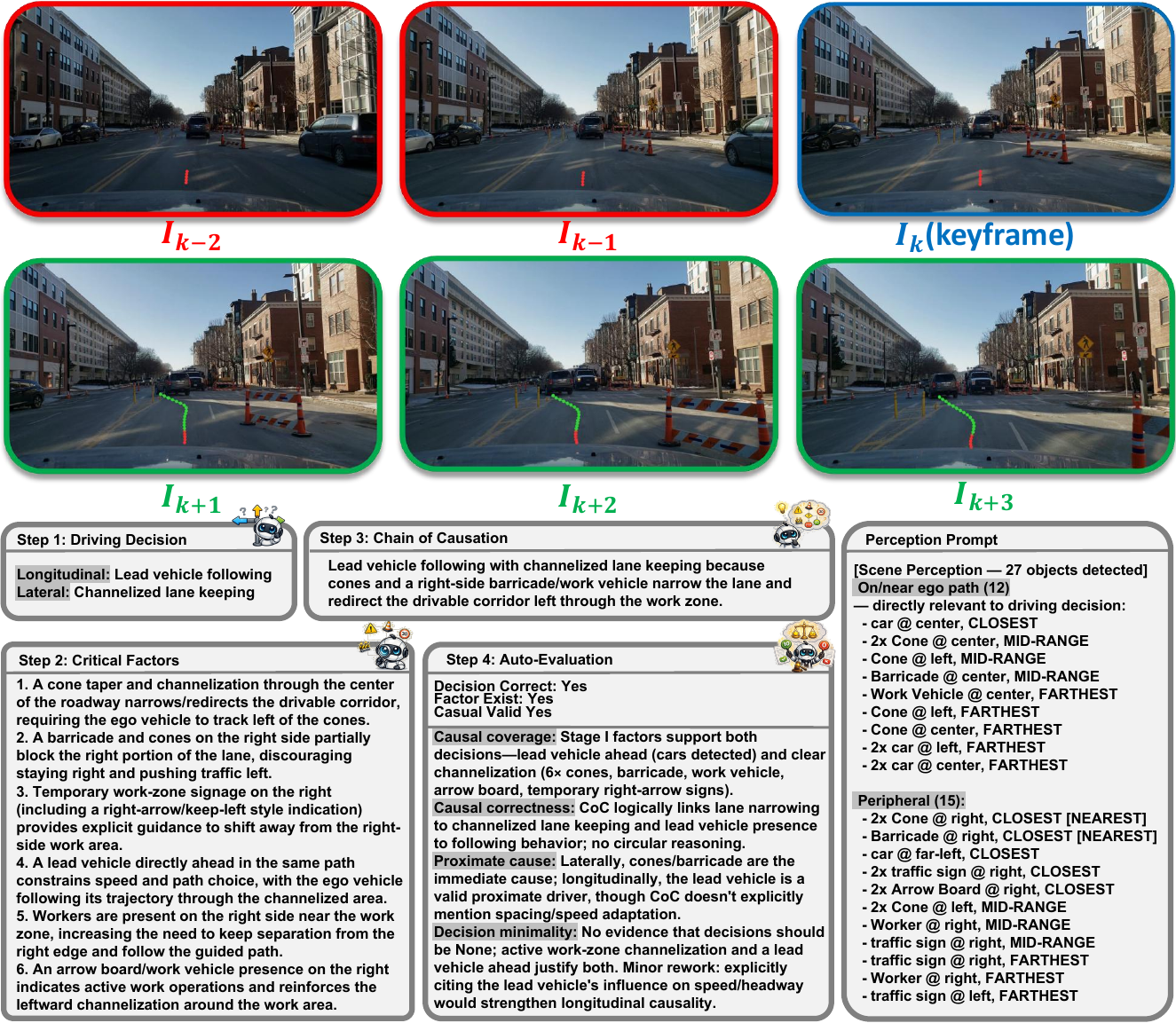}
    \caption{A complete Chain-of-Causation annotation example produced by the four-step pipeline. Perception prompt $\mathcal{P}_k$ is injected at every step to ensure that reasoning remains anchored in observable evidence.}
    \label{fig:coc-case}
\end{figure}

\begin{table*}[t]
\centering
\caption{Main results. Upper block: zero-shot VLM baselines without fine-tuning. Lower block: WorkDrive variants with GPT-5.2high labels and perception grounding (best annotation setting). \textbf{Bold} denotes the best result; \underline{underlining} denotes the second-best result.}
\label{tab:main}
\small
\setlength{\tabcolsep}{5pt}
\begin{tabular}{@{}lcccccc@{}}
\toprule
Method & ADE@5$\downarrow$ & ADE@10$\downarrow$ & ADE@15$\downarrow$ & FDE$\downarrow$ & CR (\%)$\downarrow$ & Meta-Consistency (\%)$\uparrow$ \\
\midrule
\multicolumn{7}{l}{\textit{Zero-shot baselines (no fine-tuning)}} \\
Qwen3-VL-8B-Instruct    & 3.63 & 8.69  & 18.05 & 49.83 & 19.03 & --- \\
Qwen3-VL-32B-Instruct   & 6.06 & 10.39 & 19.25 & 49.73 & 19.82 & --- \\
GPT-5.2high   & 2.13 & 6.97  & 16.09 & 47.23 & 17.43 & --- \\
Gemini 3 Pro  & 2.30 & 8.08  & 19.44 & 58.71 & 21.55 & --- \\
\midrule
\multicolumn{7}{l}{\textit{WorkDrive (GPT-5.2high + perception grounding)}} \\
Trajectory-only (SFT) & \underline{0.74} & 3.85 & 12.10 & 42.34 & 14.66 & --- \\
+ Reasoning         & 0.75 & 3.78 & 11.87 & 41.47 & 12.34 & --- \\
+ R + Meta-action   & \textbf{0.72} & \textbf{3.52} & \underline{11.01} & \underline{38.35} & \underline{11.49} & \underline{80.2} \\
+ GRPO Alignment    & \underline{0.74} & \underline{3.64} & \textbf{10.68} & \textbf{35.87} & \textbf{7.03} & \textbf{100.0} \\
\bottomrule
\end{tabular}
\end{table*}

\subsection{Reasoning SFT and Consistency Alignment}
\label{sec:alignment}

With reasoning annotations ready, Phase~III trains the VLM in two stages: supervised fine-tuning teaches the model to generate structured reasoning and trajectories, and GRPO~\cite{shao2024deepseekmath} reinforcement learning then enforces consistency between the generated reasoning and the output trajectory.

CoC annotations from Phase~II are converted into a multi-turn conversational format.
Each sample's input $x$ contains the keyframe, two preceding frames, and the observed trajectory; the target output $y$ contains three segments delimited by special tokens: \texttt{<reason>}, \texttt{<meta\_action>} representing one of the five lateral intent categories mapped from lateral decisions, and \texttt{<trajectory>} encoding 15 pixel-coordinate waypoints.
Optimization proceeds via standard cross-entropy loss over the target token sequence:
\begin{equation}
    \mathcal{L}_{\text{SFT}}(\theta) = - \sum_{i=1}^{|y|} \log P_\theta(y_i \mid x,\; y_{<i}).
    \label{eq:sft}
\end{equation}

Supervised fine-tuning injects the reasoning pattern but cannot enforce semantic consistency between the declared intent and the output trajectory.
We introduce meta-actions as a verifiable bridge: when the model correctly describes its lateral intent and the trajectory matches that intent, trajectory quality improves even without an explicit trajectory regression loss.

During reinforcement learning alignment, the policy $\pi_\theta$ samples $G{=}4$ outputs per prompt, computes group-relative advantages $A_i = R_i - \bar{R}$, and optimizes the following objective:
\begin{equation}
    \mathcal{L}_{\text{GRPO}}(\theta) = -\mathbb{E}_{o_i \sim \pi_\theta} \left[ \frac{\exp(\beta A_i)}{\sum_{j=1}^{G} \exp(\beta A_j)} \left( \log \pi_\theta(o_i) - \lambda_{\text{KL}} \, \text{KL}[\pi_\theta \| \pi_{\text{ref}}] \right) \right],
    \label{eq:grpo}
\end{equation}
where $R_i$ is the reward for the $i$-th sample (defined below), $\bar{R}$ is the group mean, and $\pi_{\text{ref}}$ is the frozen SFT checkpoint. Training hyperparameters are detailed in Appendix~C.

The reward $R_i$ for each sampled output combines a format reward with a consistency score:
\begin{equation}
    R_i = R_{\text{format}} \times R_{\text{consistency}}.
    \label{eq:reward}
\end{equation}
$R_{\text{format}} \in \{0,1\}$ checks whether the output contains the three required tokens (\texttt{<reason>}, \texttt{<meta\_action>}, \texttt{<trajectory>}) in the correct order, zeroing out the entire reward for any malformed output.
$R_{\text{consistency}}$ measures the agreement between the declared meta-action $\mathcal{M}$ and the behavioral direction derived from the mean trajectory curvature:
\begin{equation}
    \bar{\kappa} = \mathrm{mean}\left(\frac{\mathbf{v}_{i} \times \mathbf{v}_{i+1}}{|\mathbf{v}_{i}| \cdot |\mathbf{v}_{i+1}|}\right),
    \label{eq:curvature}
\end{equation}
where $\mathbf{v}_i$ denotes direction vectors between adjacent trajectory points.
In the pixel coordinate system $\bar{\kappa} > 0$ corresponds to right turns and $\bar{\kappa} < 0$ to left turns.
Specifically, $|\bar{\kappa}| < 0.03$ maps to \textit{Go straight}, $0.03 \leq |\bar{\kappa}| < 0.08$ to \textit{Steer left/right}, and $|\bar{\kappa}| \geq 0.08$ to \textit{Sharp steer left/right}, where thresholds correspond to the 25th/75th and 5th/95th percentiles of the training curvature distribution.
A graded comparison between $\hat{\mathcal{M}}$ and the declared $\mathcal{M}$ yields:
\begin{equation}
    R_{\text{consistency}} = 
    \begin{cases}
        1.0 & \text{if } \hat{\mathcal{M}} = \mathcal{M}, \\
        0.5 & \text{if } \mathrm{dir}(\hat{\mathcal{M}}) = \mathrm{dir}(\mathcal{M}) \text{ but intensity differs}, \\
        0.0 & \text{otherwise}.
    \end{cases}
    \label{eq:consistency}
\end{equation}

%
%
%

\section{Experiments}

\subsection{Experimental Setup}

\textbf{Dataset and split.}
We evaluate WorkDrive on the ROADWork dataset~\cite{ghosh2025roadwork}, the largest public work-zone benchmark.
The original dataset provides detection and trajectory annotations but defines no reasoning-oriented evaluation split.
We design a clip-level partition using 32-character hexadecimal scene IDs as the atomic unit, ensuring that all frames from a given clip reside in the same partition and eliminating frame-level data leakage.
City-stratified sampling allocates approximately 12\% of each city's clips to validation, with a minimum of one clip per city, so that every city is represented in both splits without holding out any city entirely.
The resulting split contains 801 training and 110 validation clips, yielding 14{,}331 training and 2{,}032 validation keyframe samples.

\begin{table*}[t]
\centering
\caption{Ablation study. Two factors: Label Source (GPT-5.2high vs.\ Qwen3-VL-32B-Instruct) $\times$ Perception Grounding (Yes vs.\ No). The trajectory-only baseline is shared across all conditions. \textbf{Bold} denotes the best result; \underline{underlining} denotes the second-best result.}
\label{tab:ablation}
\small
\setlength{\tabcolsep}{4pt}
\begin{tabular}{@{}lcccccccc@{}}
\toprule
Method & Perception & Label Source & ADE@5$\downarrow$ & ADE@10$\downarrow$ & ADE@15$\downarrow$ & FDE$\downarrow$ & CR (\%)$\downarrow$ & Meta-Consistency (\%)$\uparrow$ \\
\midrule
Trajectory-only & --- & --- & \underline{0.74} & 3.85 & 12.10 & 42.34 & 14.66 & --- \\
\midrule
+ Reasoning       & No  & Qwen3-VL-32B-Instruct & 0.83 & 3.99 & 12.11 & 41.81 & 13.21 & --- \\
+ Reasoning       & No  & GPT-5.2high  & \underline{0.74} & 3.78 & 11.91 & 41.82 & 12.62 & --- \\
+ Reasoning       & Yes & Qwen3-VL-32B-Instruct & 0.80 & 3.95 & 12.03 & 42.70 & 13.51 & --- \\
+ Reasoning       & Yes & GPT-5.2high  & 0.75 & 3.78 & 11.87 & 41.47 & 12.34 & --- \\
\midrule
+ R + Meta-action & No  & Qwen3-VL-32B-Instruct & 0.82 & 3.86 & 11.66 & 39.80 & 8.90  & 78.9 \\
+ R + Meta-action & No  & GPT-5.2high  & 0.75 & 3.68 & 11.37 & 39.39 & 10.74 & 79.6 \\
+ R + Meta-action & Yes & Qwen3-VL-32B-Instruct & 0.81 & 3.82 & 11.58 & 39.79 & 8.39  & 79.2 \\
+ R + Meta-action & Yes & GPT-5.2high  & \textbf{0.72} & \textbf{3.52} & 11.01 & 38.35 & 11.49 & 80.2 \\
\midrule
+ GRPO            & No  & Qwen3-VL-32B-Instruct & 0.80 & 3.70 & 11.03 & 37.00 & 7.99  & 98.1 \\
+ GRPO            & No  & GPT-5.2high  & \textbf{0.72} & 3.70 & 11.08 & \underline{36.32} & 7.82  & \underline{99.3} \\
+ GRPO            & Yes & Qwen3-VL-32B-Instruct & 0.79 & 3.68 & \underline{10.81} & 36.35 & \underline{7.43} & \textbf{100.0} \\
+ GRPO            & Yes & GPT-5.2high  & \underline{0.74} & \underline{3.64} & \textbf{10.68} & \textbf{35.87} & \textbf{7.03} & \textbf{100.0} \\
\bottomrule
\end{tabular}
\end{table*}

\textbf{Metrics.}
Evaluation spans two complementary dimensions: trajectory accuracy and reasoning-behavior consistency.
Following prior work~\cite{sima2024drivelm,hwang2024emma,liao2025work}, we adopt three standard open-loop planning metrics.
Average Displacement Error at horizon $k$ (ADE@$k$) measures the mean $\ell_2$ distance between the first $k$ predicted waypoints and their ground-truth counterparts:
\begin{equation}
    \text{ADE@}k = \frac{1}{k}\sum_{t=1}^{k}\|\hat{\mathbf{p}}_t - \mathbf{p}_t\|_2,
    \label{eq:ade}
\end{equation}
where $\hat{\mathbf{p}}_t, \mathbf{p}_t \in \mathbb{R}^2$ are the predicted and ground-truth positions at step $t$.
We report $k \in \{5, 10, 15\}$; ADE@15 is the primary metric.
Final Displacement Error (FDE) measures the displacement at the last predicted waypoint:
\begin{equation}
    \text{FDE} = \|\hat{\mathbf{p}}_T - \mathbf{p}_T\|_2,
    \label{eq:fde}
\end{equation}
where $T{=}15$ is the prediction horizon.
Collision Rate (CR) reports the fraction of predicted trajectories that intersect obstacles:
\begin{equation}
    \text{CR} = \frac{1}{N}\sum_{i=1}^{N}\mathbb{1}\!\left[\hat{Y}_i \text{ collides}\right],
    \label{eq:cr}
\end{equation}
where $N$ is the total number of samples and $\mathbb{1}[\cdot]$ is the indicator function.
Because ROADWork provides annotations in pixel coordinates rather than BEV, all displacement metrics are measured in pixels.
For all three metrics, lower is better.
For reasoning-behavior consistency, we report Meta-Consistency, the agreement rate between the model's declared lateral meta-action and the curvature-derived direction of the predicted trajectory (Eq.~\ref{eq:consistency}); this metric also serves as the sole reward signal during GRPO alignment (\S\ref{sec:alignment}).

\textbf{Implementation details.}
All WorkDrive variants use Qwen3-VL-8B-Instruct~\cite{bai2025qwen3} as the base model.
SFT employs LoRA~\cite{hu2022lora} (rank\,=\,64, $\alpha$\,=\,128) and runs for 10 epochs with a learning rate of $2 \times 10^{-5}$ (cosine schedule, warmup ratio 0.1).
GRPO~\cite{shao2024deepseekmath} performs full-parameter training from the merged SFT checkpoint, with a learning rate of $1 \times 10^{-6}$ for one epoch, group sampling size $G{=}4$, no KL regularization, and a single reward: meta-consistency.
Zero-shot baselines receive the same prompt format as WorkDrive and generate 15-step trajectories without fine-tuning.
ROADWork provides trajectory annotations in 2D pixel coordinates rather than the BEV or 3D world coordinates used by nuScenes and Waymo.
The most closely related method, REACT-Drive~\cite{liao2025work}, has not released its code or model weights, precluding direct reproduction.
More broadly, existing open-source reasoning-augmented driving methods are designed for BEV coordinate systems and do not support pixel-coordinate trajectory prediction.
We therefore evaluate against zero-shot VLM baselines under the same coordinate convention, which isolates the contribution of perception-grounded reasoning and consistency alignment.

\subsection{Main Results}
\label{sec:main-results}

Table~\ref{tab:main} compares zero-shot baselines with WorkDrive under the best annotation setting (GPT-5.2high~\cite{openai2025gpt52} labels with perception grounding); effects of individual factors are isolated in the ablation study (\S\ref{sec:ablation}).

\textbf{Work zones pose a severe OOD challenge.}
Even the strongest zero-shot baseline, GPT-5.2high~\cite{openai2025gpt52}, reaches only 16.09 ADE@15, 47.23 FDE, and a 17.43\% collision rate.
Qwen3-VL-32B-Instruct~\cite{bai2025qwen3} further underperforms its 8B counterpart (19.25 vs.\ 18.05 ADE@15), confirming that parameter scaling alone does not help when the model lacks the causal knowledge to connect work-zone objects to planning decisions.

\textbf{Reasoning augmentation progressively improves trajectory quality.}
From the trajectory-only SFT baseline at 12.10 ADE@15, adding CoC reasoning reduces ADE@15 to 11.87 and lowers the collision rate from 14.66\% to 12.34\%.
Incorporating the meta-action brings a larger gain---ADE@15 drops to 11.01 ($-9.0\%$) and FDE from 42.34 to 38.35 ($-9.4\%$)---while yielding the first measurable Meta-Consistency of 80.2\%, as the model begins to explicitly align its declared lateral intent with the predicted trajectory.

\textbf{Consistency alignment drives the final performance gain.}
GRPO further reduces ADE@15 to 10.68, FDE to 35.87, and CR to 7.03\%, yielding additional improvements of $-3.0\%$, $-6.5\%$, and $-38.8\%$, respectively, over the +R+Meta baseline while raising Meta-Consistency from 80.2\% to 100.0\%.
Crucially, meta-action consistency is the \emph{sole} reward; no explicit trajectory regression signal is used, yet the long-horizon trajectory metrics (ADE@15, FDE, and CR) continue to improve.
This validates the core hypothesis of \S\ref{sec:alignment}: when reasoning accurately describes the lateral intent and the trajectory faithfully executes it, consistency alone constitutes an effective alignment signal.
WorkDrive's single-reward design thus demonstrates that consistency alone suffices, without the multi-reward schemes used in prior work such as AR1~\cite{wang2025alpamayo}.

\begin{figure}[t]
    \centering
    \includegraphics[width=0.80\linewidth]{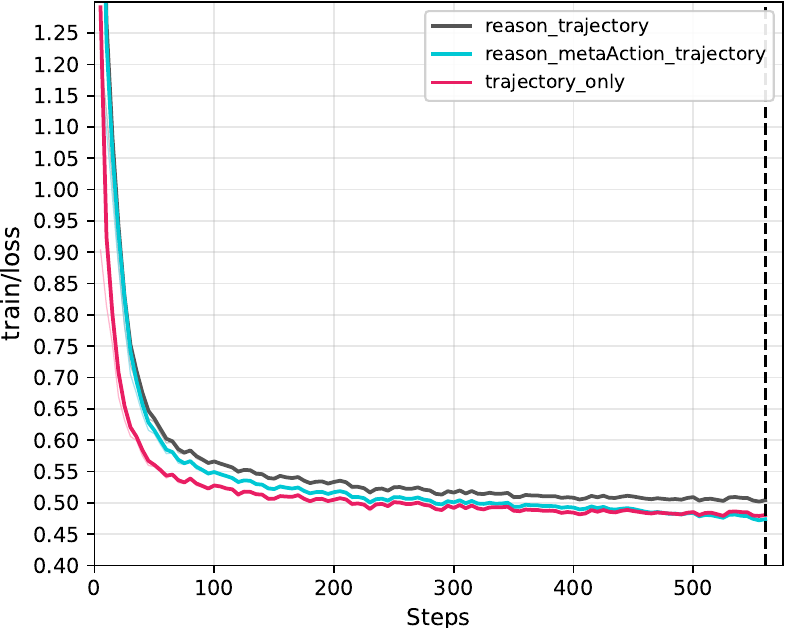}
    \caption{SFT training loss curves across three configurations.}
    \label{fig:sft-loss}
    \vspace{-0.1in}
\end{figure}

\subsection{Ablation Studies}
\label{sec:ablation}


Table~\ref{tab:ablation} isolates two factors, label source and perception grounding, using the trajectory-only baseline as a common reference.

\textbf{Perception grounding consistently improves trajectory accuracy.}
Under GPT labels, grounding reduces ADE@15 from 11.37 to 11.01 at the SFT stage, and a similar pattern holds for Qwen labels (11.66\,$\to$\,11.58).
This benefit amplifies after GRPO: the GPT + perception configuration reaches 10.68 ADE@15, a 3.6\% reduction over its ungrounded counterpart at 11.08.
The amplification arises because on-policy sampling in RL cascades annotation quality differences; perception grounding provides a stronger SFT initialization, enabling GRPO to explore from a higher-quality starting point.
This result confirms that redirecting annotation attention to domain-specific work-zone elements during data construction propagates through the entire training pipeline.

\textbf{GRPO narrows the label quality gap.}
At the SFT stage with perception, GPT labels outperform Qwen labels by 4.9\% in ADE@15 (11.01 vs.\ 11.58).
After GRPO, this gap narrows to 1.2\% (10.68 vs.\ 10.81) and both configurations reach 100.0\% Meta-Consistency.
This demonstrates the label-noise tolerance of consistency-based RL: as long as the reward provides a clear gradient signal, the model compensates for annotation noise through online exploration. In practice, open-source annotations calibrated through GRPO can therefore substitute for costly closed-source labels.

\textbf{Method stages exhibit a consistent incremental pattern.}
Across all four annotation conditions, the three stages follow a stable progression.
The +Reasoning stage yields modest ADE improvement but substantial CR reduction, from 14.66\% to 12--13\%, as reasoning supervision primarily constrains extreme trajectory deviations.
The +Meta-action stage produces a more pronounced ADE@15 decrease of 3.7--7.2\%, as verifiable lateral intent anchors strengthen the connection between reasoning and trajectory.
The +GRPO stage raises Meta-Consistency to 98--100\% and improves ADE, FDE, and CR uniformly, confirming that consistency alignment is robust to both label source and perception configuration.

\begin{figure}[t]
    \centering
    \includegraphics[width=0.84\linewidth]{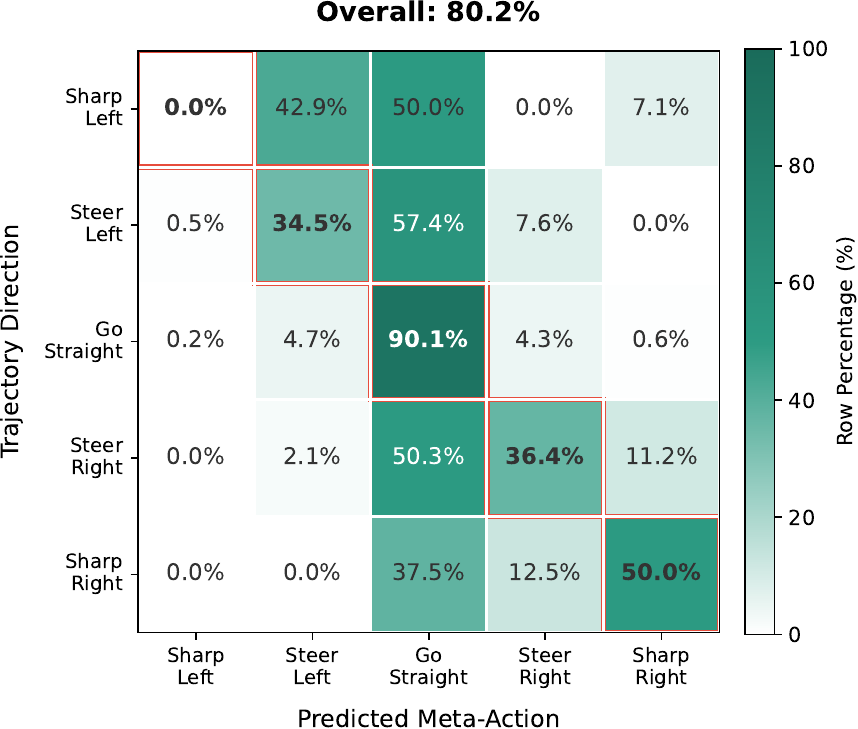}
    \caption{Meta-action confusion matrix of the SFT model.}
    \label{fig:meta-confusion}
\end{figure}

\subsection{Analysis}

\textbf{SFT training dynamics.}
Figure~\ref{fig:sft-loss} compares training loss curves for three SFT configurations: trajectory-only, +Reasoning, and +R+Meta-action.
Adding reasoning text raises the absolute loss because the target now includes language tokens, but does not slow the convergence of trajectory tokens.
The +R+Meta-action configuration converges fastest, because meta-action gradients regularize the joint reasoning-trajectory optimization.
All three configurations plateau between epochs 5 and 7 without overfitting.

\textbf{Meta-action confusion before and after GRPO.}
Figure~\ref{fig:meta-confusion} shows the meta-action confusion matrix of the SFT model prior to GRPO.
The dominant error mode is a mismatch between \textit{Go straight} trajectories and \textit{Steer left/right} declarations, which accounts for most of the remaining 19.8\% inconsistency.
After GRPO, the consistency reward eliminates these misalignments and drives Meta-Consistency to 100.0\%.
This confirms that token-level cross-entropy cannot enforce semantic coherence across output modalities, whereas the consistency reward directly penalizes intent--behavior discrepancies.

\begin{figure}[t]
    \centering
    \includegraphics[width=\linewidth]{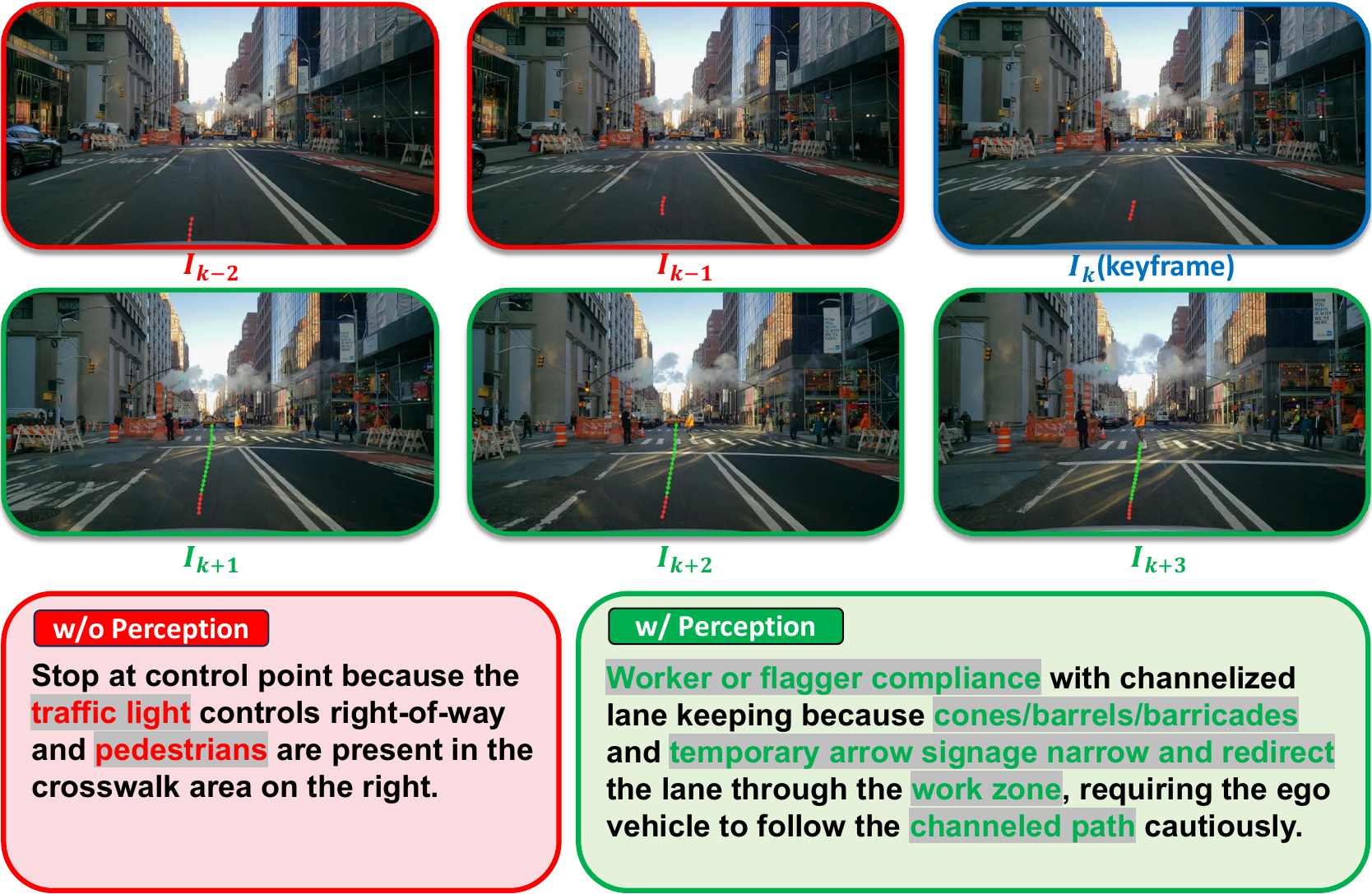}
    \caption{Qualitative comparison of CoC reasoning with and without perception grounding on the same keyframe.}
    \label{fig:perception-case}
    \vspace{-0.1in}
\end{figure}

\textbf{Qualitative comparison of perception grounding.}
Figure~\ref{fig:perception-case} contrasts CoC reasoning for the same keyframe with and without perception grounding.
Without perception grounding, the annotator defaults to generic descriptions such as ``a construction zone ahead'' without referencing specific objects or their spatial arrangement.
With perception grounding, the reasoning directly references structured facts from the perception pipeline, citing concrete object counts, positions, and depth levels, and anchors the lateral decision to the work-zone elements that actually define the drivable corridor.
This qualitative difference directly illustrates the attention-redirecting effect described in \S\ref{sec:perception}: structured perception information shifts the annotator's focus from generic scene labels to the domain-specific elements that govern the driving decision.

\textbf{Human evaluation of annotation quality.}
\label{sec:human-eval}
To validate the Step~4 automated quality assessment, an annotator with experience in autonomous driving independently scored 100 randomly sampled CoC annotations from the training set using the same four criteria on a 0--10 integer scale.
Table~\ref{tab:human-eval} compares human scores against three candidate judge models.
GPT-5.2high~\cite{openai2025gpt52} achieves the closest alignment with human ratings: its mean absolute error is only 0.62 points and its pass-rate agreement reaches 92\%, substantially outperforming Gemini 3 Pro~\cite{google2025gemini3} and Qwen3-VL-32B-Instruct~\cite{bai2025qwen3}.

\begin{table}[t]
\centering
\caption{Human--automated-judge agreement on 100 CoC annotations sampled from the training set. MAE: mean absolute error between human and automated-judge scores. Pass Agreement: percentage of samples for which human and automated judges agree on the pass/fail decision ($s \geq 8$).}
\label{tab:human-eval}
\small
\setlength{\tabcolsep}{5pt}
\begin{tabular}{@{}lccc@{}}
\toprule
Judge Model & Mean Score & MAE (pts)$\downarrow$ & Pass Agreement (\%)$\uparrow$ \\
\midrule
Human   & $8.41 \pm 1.07$ & ---  & --- \\
\midrule
GPT-5.2high         & $8.53 \pm 0.94$ & \textbf{0.62} & \textbf{92} \\
Gemini 3 Pro        & $8.71 \pm 0.81$ & 0.89 & 85 \\
Qwen3-VL-32B-Instruct        & $9.17 \pm 0.58$ & 1.18 & 81 \\
\bottomrule
\end{tabular}
\end{table}

%
\section{Conclusion}

We present WorkDrive, a framework that addresses a data gap exposed when vision-language models encounter work zones: VLMs can detect temporary traffic controls such as cones and barriers, yet no causal-reasoning annotations exist to teach how these devices reconfigure drivable corridors and necessitate lateral avoidance. WorkDrive fills this gap bottom-up through automated perception, using an extraction pipeline to produce structured scene facts, a Chain-of-Causation annotation pipeline to convert these facts into causally grounded reasoning labels, and Group Relative Policy Optimization to align reasoning with planning via a single reward: consistency between lateral meta-actions and trajectory curvature. On ROADWork, perception-grounded reasoning reduces ADE by 9.0\% and consistency-based GRPO yields an additional 3.0\%, achieving progressive improvement over the trajectory-only baseline while halving the collision rate, with no explicit trajectory reward at the alignment stage. This result demonstrates that, when reasoning is grounded in perception, behavioral consistency between reasoning and planning can substitute for explicit trajectory supervision under distributional shift. We will publicly release the code and reasoning dataset to facilitate research on extending perception-grounded reasoning to other long-tail driving scenarios.

\bibliographystyle{ACM-Reference-Format}
\bibliography{samples/sample-base}

@inproceedings{hu2023planning,
  title={Planning-oriented autonomous driving},
  author={Hu, Yihan and Yang, Jiazhi and Chen, Li and Li, Keyu and Sima, Chonghao and Zhu, Xizhou and Chai, Siqi and Du, Senyao and Lin, Tianwei and Wang, Wenhai and Lu, Lewei and Jia, Xiaosong and Liu, Qiang and Dai, Jifeng and Qiao, Yu and Li, Hongyang},
  booktitle={Proceedings of the IEEE/CVF conference on computer vision and pattern recognition},
  pages={17853--17862},
  year={2023}
}

@article{tian2024drivevlm,
  title={Drivevlm: The convergence of autonomous driving and large vision-language models},
  author={Tian, Xiaoyu and Gu, Junru and Li, Bailin and Liu, Yicheng and Wang, Yang and Zhao, Zhiyong and Zhan, Kun and Jia, Peng and Lang, Xianpeng and Zhao, Hang},
  journal={arXiv preprint arXiv:2402.12289},
  year={2024}
}

@article{hwang2024emma,
  title={Emma: End-to-end multimodal model for autonomous driving},
  author={Jyh-Jing Hwang and Runsheng Xu and Hubert Lin and Wei-Chih Hung and Jingwei Ji and Kristy Choi and Di Huang and Tong He and Paul Covington and Benjamin Sapp and Yin Zhou and James Guo and Dragomir Anguelov and Mingxing Tan},
  journal={arXiv preprint arXiv:2410.23262},
  year={2024}
}

@article{jiang2024senna,
  title={Senna: Bridging large vision-language models and end-to-end autonomous driving},
  author={Jiang, Bo and Chen, Shaoyu and Liao, Bencheng and Zhang, Xingyu and Yin, Wei and Zhang, Qian and Huang, Chang and Liu, Wenyu and Wang, Xinggang},
  journal={arXiv preprint arXiv:2410.22313},
  year={2024}
}

@inproceedings{zhou2026opendrivevla,
  title={Opendrivevla: Towards end-to-end autonomous driving with large vision language action model},
  author={Zhou, Xingcheng and Han, Xuyuan and Yang, Feng and Ma, Yunpu and Tresp, Volker and Knoll, Alois},
  booktitle={Proceedings of the AAAI Conference on Artificial Intelligence},
  volume={40},
  number={16},
  pages={13782--13790},
  year={2026}
}

@inproceedings{fu2025orion,
  title={Orion: A holistic end-to-end autonomous driving framework by vision-language instructed action generation},
  author={Fu, Haoyu and Zhang, Diankun and Zhao, Zongchuang and Cui, Jianfeng and Liang, Dingkang and Zhang, Chong and Zhang, Dingyuan and Xie, Hongwei and Wang, Bing and Bai, Xiang},
  booktitle={Proceedings of the IEEE/CVF International Conference on Computer Vision},
  pages={24823--24834},
  year={2025}
}

@article{jiang2025diffvla,
  title={Diffvla: Vision-language guided diffusion planning for autonomous driving},
  author={Anqing Jiang and Yu Gao and Zhigang Sun and Yiru Wang and Jijun Wang and Jinghao Chai and Qian Cao and Yuweng Heng and Hao Jiang and Yunda Dong and Zongzheng Zhang and Xianda Guo and Hao Sun and Hao Zhao},
  journal={arXiv preprint arXiv:2505.19381},
  year={2025}
}

@article{peng2025colavla,
  title={ColaVLA: Leveraging Cognitive Latent Reasoning for Hierarchical Parallel Trajectory Planning in Autonomous Driving},
  author={Peng, Qihang and Chen, Xuesong and Yang, Chenye and Shi, Shaoshuai and Li, Hongsheng},
  journal={arXiv preprint arXiv:2512.22939},
  year={2025}
}

@article{wang2024drivecot,
  title={Drivecot: Integrating chain-of-thought reasoning with end-to-end driving},
  author={Wang, Tianqi and Xie, Enze and Chu, Ruihang and Li, Zhenguo and Luo, Ping},
  journal={arXiv preprint arXiv:2403.16996},
  year={2024}
}

@inproceedings{sima2024drivelm,
  title={Drivelm: Driving with graph visual question answering},
  author={Sima, Chonghao and Renz, Katrin and Chitta, Kashyap and Chen, Li and Zhang, Hanxue and Xie, Chengen and Bei{\ss}wenger, Jens and Luo, Ping and Geiger, Andreas and Li, Hongyang},
  booktitle={European conference on computer vision},
  pages={256--274},
  year={2024},
  organization={Springer}
}

@article{wang2025cot4ad,
  title={CoT4AD: A Vision-Language-Action Model with Explicit Chain-of-Thought Reasoning for Autonomous Driving},
  author={Wang, Zhaohui and Yu, Tengbo and Tang, Hao},
  journal={arXiv preprint arXiv:2511.22532},
  year={2025}
}

@article{wang2025alpamayo,
  title={Alpamayo-r1: Bridging reasoning and action prediction for generalizable autonomous driving in the long tail},
  author={Yan Wang and Wenjie Luo and Junjie Bai and Yulong Cao and Tong Che and Ke Chen and Yuxiao Chen and Jenna Diamond and Yifan Ding and Wenhao Ding and Liang Feng and Greg Heinrich and Jack Huang and Peter Karkus and Boyi Li and Pinyi Li and Tsung-Yi Lin and Dongran Liu and Ming-Yu Liu and Langechuan Liu and Zhijian Liu and Jason Lu and Yunxiang Mao and Pavlo Molchanov and Lindsey Pavao and Zhenghao Peng and Mike Ranzinger and Ed Schmerling and Shida Shen and Yunfei Shi and Sarah Tariq and Ran Tian and Tilman Wekel and Xinshuo Weng and Tianjun Xiao and Eric Yang and Xiaodong Yang and Yurong You and Xiaohui Zeng and Wenyuan Zhang and Boris Ivanovic and Marco Pavone},
  journal={arXiv preprint arXiv:2511.00088},
  year={2025}
}

@article{jiang2025alphadrive,
  title={Alphadrive: Unleashing the power of vlms in autonomous driving via reinforcement learning and reasoning},
  author={Jiang, Bo and Chen, Shaoyu and Zhang, Qian and Liu, Wenyu and Wang, Xinggang},
  journal={arXiv preprint arXiv:2503.07608},
  year={2025}
}

@inproceedings{li2026drive,
  title={Drive-r1: Bridging reasoning and planning in vlms for autonomous driving with reinforcement learning},
  author={Li, Yue and Tian, Meng and Zhu, Dechang and Zhu, Jiangtong and Lin, Zhenyu and Xiong, Zhiwei and Zhao, Xinhai},
  booktitle={Proceedings of the AAAI Conference on Artificial Intelligence},
  volume={40},
  number={8},
  pages={6708--6716},
  year={2026}
}

@article{zhang2025omnidrive,
  title={OmniDrive-R1: Reinforcement-driven Interleaved Multi-modal Chain-of-Thought for Trustworthy Vision-Language Autonomous Driving},
  author={Zhang, Zhenguo and Zheng, Haohan and Wang, Yishen and Xu, Le and Deng, Tianchen and Chen, Xuefeng and Chen, Qu and Zhang, Bo and Huang, Wuxiong},
  journal={arXiv preprint arXiv:2512.14044},
  year={2025}
}

@article{zheng2025driveagent,
  title={DriveAgent-R1: Advancing VLM-based Autonomous Driving with Active Perception and Hybrid Thinking},
  author={Zheng, Weicheng and Mao, Xiaofei and Ye, Nanfei and Li, Pengxiang and Zhan, Kun and Lang, Xianpeng and Zhao, Hang},
  journal={arXiv preprint arXiv:2507.20879},
  year={2025}
}

@inproceedings{ghosh2025roadwork,
  title={Roadwork: A dataset and benchmark for learning to recognize, observe, analyze and drive through work zones},
  author={Ghosh, Anurag and Zheng, Shen and Tamburo, Robert and Vuong, Khiem and Alvarez-Padilla, Juan and Zhu, Hailiang and Cardei, Michael and Dunn, Nicholas and Mertz, Christoph and Narasimhan, Srinivasa G},
  booktitle={Proceedings of the IEEE/CVF International Conference on Computer Vision},
  pages={6132--6142},
  year={2025}
}

@article{liao2025work,
  title={Work zones challenge vlm trajectory planning: Toward mitigation and robust autonomous driving},
  author={Liao, Yifan and Sun, Zhen and Qiu, Xiaoyun and Zhao, Zixiao and Tang, Wenbing and He, Xinlei and Zheng, Xinhu and Zhang, Tianwei and Huang, Xinyi and Han, Xingshuo},
  journal={arXiv preprint arXiv:2510.02803},
  year={2025}
}

@techreport{penndot2019ads,
  author       = {{Pennsylvania Department of Transportation}},
  title        = {Safe Integration of Automated Vehicles into Work Zones},
  institution  = {U.S. Department of Transportation},
  year         = {2019},
  type         = {Technical Report},
  note         = {Accessed: 2026-03-29},
}

@article{shao2024deepseekmath,
  title={Deepseekmath: Pushing the limits of mathematical reasoning in open language models},
  author={Shao, Zhihong and Wang, Peiyi and Zhu, Qihao and Xu, Runxin and Song, Junxiao and Bi, Xiao and Zhang, Haowei and Zhang, Mingchuan and Li, YK and Wu, Yang and Daya, Guo},
  journal={arXiv preprint arXiv:2402.03300},
  year={2024}
}

@inproceedings{caesar2020nuscenes,
  title={nuscenes: A multimodal dataset for autonomous driving},
  author={Caesar, Holger and Bankiti, Varun and Lang, Alex H and Vora, Sourabh and Liong, Venice Erin and Xu, Qiang and Krishnan, Anush and Pan, Yu and Baldan, Giancarlo and Beijbom, Oscar},
  booktitle={Proceedings of the IEEE/CVF conference on computer vision and pattern recognition},
  pages={11621--11631},
  year={2020}
}

@inproceedings{sun2020scalability,
  title={Scalability in perception for autonomous driving: Waymo open dataset},
  author={Pei Sun and Henrik Kretzschmar and Xerxes Dotiwalla and Aurelien Chouard and Vijaysai Patnaik and Paul Tsui and James Guo and Yin Zhou and Yuning Chai and Benjamin Caine and Vijay Vasudevan and Wei Han and Jiquan Ngiam and Hang Zhao and Aleksei Timofeev and Scott Ettinger and Maxim Krivokon and Amy Gao and Aditya Joshi and Sheng Zhao and Shuyang Cheng and Yu Zhang and Jonathon Shlens and Zhifeng Chen and Dragomir Anguelov},
  booktitle={Proceedings of the IEEE/CVF conference on computer vision and pattern recognition},
  pages={2446--2454},
  year={2020}
}

@article{xu2024drivegpt4,
  title={Drivegpt4: Interpretable end-to-end autonomous driving via large language model},
  author={Xu, Zhenhua and Zhang, Yujia and Xie, Enze and Zhao, Zhen and Guo, Yong and Wong, Kwan-Yee K and Li, Zhenguo and Zhao, Hengshuang},
  journal={IEEE Robotics and Automation Letters},
  volume={9},
  number={10},
  pages={8186--8193},
  year={2024},
  publisher={IEEE}
}

@article{li2025recogdrive,
  title={Recogdrive: A reinforced cognitive framework for end-to-end autonomous driving},
  author={Yongkang Li and Kaixin Xiong and Xiangyu Guo and Fang Li and Sixu Yan and Gangwei Xu and Lijun Zhou and Long Chen and Haiyang Sun and Bing Wang and Kun Ma and Guang Chen and Hangjun Ye and Wenyu Liu and Xinggang Wang},
  journal={arXiv preprint arXiv:2506.08052},
  year={2025}
}

@article{zhou2025autovla,
  title={Autovla: A vision-language-action model for end-to-end autonomous driving with adaptive reasoning and reinforcement fine-tuning},
  author={Zhou, Zewei and Cai, Tianhui and Zhao, Seth Z and Zhang, Yun and Huang, Zhiyu and Zhou, Bolei and Ma, Jiaqi},
  journal={arXiv preprint arXiv:2506.13757},
  year={2025}
}

@article{luo2025adathinkdrive,
  title={Adathinkdrive: Adaptive thinking via reinforcement learning for autonomous driving},
  author={Yuechen Luo and Fang Li and Shaoqing Xu and Zhiyi Lai and Lei Yang and Qimao Chen and Ziang Luo and Zixun Xie and Shengyin Jiang and Jiaxin Liu and Long Chen and Bing Wang and Zhi-xin Yang},
  journal={arXiv preprint arXiv:2509.13769},
  year={2025}
}

@article{zhang2025openread,
  title={OpenREAD: Reinforced Open-Ended Reasoning for End-to-End Autonomous Driving with LLM-as-Critic},
  author={Zhang, Songyan and Huang, Wenhui and Chen, Zhan and Collister, Chua Jiahao and Huang, Qihang and Lv, Chen},
  journal={arXiv preprint arXiv:2512.01830},
  year={2025}
}

@manual{mutcd,
  title        = {Manual on Uniform Traffic Control Devices for Streets and Highways},
  author       = {{Federal Highway Administration}},
  year         = {2009},
  organization = {U.S. Department of Transportation},
  note         = {2009 Edition with Revisions 1 and 2},
  url          = {https://mutcd.fhwa.dot.gov/}
}

@misc{tesla2017,
  author       = {{Autoweek}},
  title        = {Tesla Model {S} Autopilot Strikes Again: {Dallas} Crash},
  year         = {2017},
  howpublished = {\url{https://www.autoweek.com/news/technology/a1816086/tesla-model-s-autopilot-strikes-again-dallas-crash}},
  note         = {News article; accessed 2026-03-29}
}

@misc{cruise2023,
  author       = {{Los Angeles Times}},
  title        = {Robot Car Collides with Fire Truck in {San Francisco}},
  year         = {2023},
  howpublished = {\url{https://www.latimes.com/california/story/2023-08-19/robot-car-collides-with-fire-truck}},
  note         = {News article; accessed 2026-03-29}
}

@inproceedings{gumpp2009recognition,
  title={Recognition and tracking of temporary lanes in motorway construction sites},
  author={Gumpp, Thomas and Nienhuser, Dennis and Liebig, Rebecca and Zollner, J Marius},
  booktitle={2009 IEEE Intelligent Vehicles Symposium},
  pages={305--310},
  year={2009},
  organization={IEEE}
}

@inproceedings{mathibela2013roadwork,
  title={A roadwork scene signature based on the opponent colour model},
  author={Mathibela, Bonolo and Posner, Ingmar and Newman, Paul},
  booktitle={2013 IEEE/RSJ International Conference on Intelligent Robots and Systems},
  pages={4394--4400},
  year={2013},
  organization={IEEE}
}

@inproceedings{li2022coda,
  title={Coda: A real-world road corner case dataset for object detection in autonomous driving},
  author={Kaican Li and Kai Chen and Haoyu Wang and Lanqing Hong and Chaoqiang Ye and Jianhua Han and Yukuai Chen and Wei Zhang and Chunjing Xu and Dit-Yan Yeung and Xiaodan Liang and Zhenguo Li and Hang Xu},
  booktitle={European Conference on Computer Vision},
  pages={406--423},
  year={2022},
  organization={Springer}
}

@inproceedings{kim2024rosa,
  title={RoSA Dataset: Road Construct Zone Segmentation for Autonomous Driving},
  author={Kim, Jinwoo and An, Kyounghwan and Lee, Donghwan},
  booktitle={European Conference on Computer Vision},
  pages={322--338},
  year={2024},
  organization={Springer}
}

@inproceedings{sural2026workzone3d,
  title={WorkZone3D: A Multimodal Dataset for 3D Work Zone Perception in Autonomous Driving},
  author={Sural, Shounak and Sahu, Nishad and Rajkumar, Ragunathan},
  booktitle={Proceedings of the IEEE/CVF Winter Conference on Applications of Computer Vision},
  pages={3972--3981},
  year={2026}
}

@inproceedings{sahu2025towards,
  title={Towards the Safe Operation of Autonomous Vehicles in Work Zones},
  author={Sahu, Nishad and Su, Gregory T and Sural, Shounak and Brennan, Sean and Rajkumar, Ragunathan Raj},
  booktitle={2025 IEEE Intelligent Vehicles Symposium (IV)},
  pages={2380--2387},
  year={2025},
  organization={IEEE}
}

@inproceedings{hu2022lora,
  title={LoRA: Low-Rank Adaptation of Large Language Models},
  author={Hu, Edward J. and Shen, Yelong and Wallis, Phillip and Allen-Zhu, Zeyuan and Li, Yuanzhi and Wang, Shean and Wang, Lu and Chen, Weizhu},
  booktitle={International Conference on Learning Representations},
  year={2022},
  url={https://openreview.net/forum?id=nZeVKeeFYf9}
}

@article{lin2025depth,
  title={Depth Anything 3: Recovering the Visual Space from Any Views},
  author={Lin, Haotong and Chen, Sili and Liew, Jun Hao and Chen, Donny Y. and Li, Zhenyu and Shi, Guang and Feng, Jiashi and Kang, Bingyi},
  journal={arXiv preprint arXiv:2511.10647},
  year={2025},
  url={https://arxiv.org/abs/2511.10647}
}

@article{jiang2025detect,
  title={Detect Anything via Next Point Prediction},
  author={Jiang, Qing and Huo, Junan and Chen, Xingyu and Xiong, Yuda and Zeng, Zhaoyang and Chen, Yihao and Ren, Tianhe and Yu, Junzhi and Zhang, Lei},
  journal={arXiv preprint arXiv:2510.12798},
  year={2025},
  url={https://arxiv.org/abs/2510.12798}
}

@article{carion2025sam,
  title={{SAM 3}: Segment Anything with Concepts},
  author={Carion, Nicolas and Gustafson, Laura and Hu, Yuan-Ting and Debnath, Shoubhik and Hu, Ronghang and Suris, Didac and Ryali, Chaitanya and others},
  journal={arXiv preprint arXiv:2511.16719},
  year={2025},
  url={https://arxiv.org/abs/2511.16719}
}

@article{bai2025qwen3,
  title={{Qwen3-VL} Technical Report},
  author={Bai, Shuai and Cai, Yuxuan and Chen, Ruizhe and Chen, Keqin and Chen, Xionghui and Cheng, Zesen and Deng, Lianghao and others},
  journal={arXiv preprint arXiv:2511.21631},
  year={2025},
  url={https://arxiv.org/abs/2511.21631}
}

@misc{openai2025gpt52,
  author={{OpenAI}},
  title={Introducing {GPT-5.2}},
  year={2025},
  url={https://openai.com/index/introducing-gpt-5-2/},
  note={Accessed: 2026-07-14}
}

@misc{google2025gemini3,
  author={{Google}},
  title={A New Era of Intelligence with {Gemini 3}},
  year={2025},
  url={https://blog.google/products-and-platforms/products/gemini/gemini-3/},
  note={Accessed: 2026-07-14}
}

\end{document}